%% file: main.tex
\documentclass[twoside,11pt]{article}

\usepackage{jmlr2e}
\usepackage[utf8]{inputenc} 
\usepackage[T1]{fontenc}    
\usepackage{hyperref}       
\usepackage{url}            
\usepackage{booktabs}       
\usepackage{amsfonts}       
\usepackage{nicefrac}       
\usepackage{microtype}      
\usepackage{xcolor}         
\usepackage{enumerate}
\usepackage{enumitem}
\usepackage{graphicx}

\usepackage{amsmath}
\usepackage{bm}
\usepackage{mwe}
\usepackage{color, colortbl}
\definecolor{lightGray}{gray}{0.1}
\definecolor{Gray}{gray}{0.9}
\definecolor{LightCyan}{rgb}{0.88,1,1}
\usepackage[T1]{fontenc}
\usepackage[font=small,labelfont=bf,tableposition=top]{caption}
\usepackage{wrapfig}
\usepackage{multirow}
\DeclareCaptionLabelFormat{andtable}{#1~#2  \&  \tablename~\thetable}
\usepackage{caption}
\usepackage{algorithm}
\usepackage{algpseudocode}
\algnewcommand\algorithmicforeach{\textbf{for each}}
\algdef{S}[FOR]{ForEach}[1]{\algorithmicforeach\ #1\ \algorithmicdo}
\usepackage{array}
\usepackage{eqparbox}

\usepackage{etoolbox}  

\jmlrheading{1}{2000}{1-48}{4/00}{10/00}{meila00a}{Marina Meil\u{a} and Michael I. Jordan}

\ShortHeadings{Bayesian Importance of Features}{Adamczewski, Harder and Park}
\firstpageno{1}

\input{notation}

\begin{document}

\title{Bayesian Importance of Features (BIF)}

\author{\name Kamil Adamczewski \email kamil.m.adamczewski@gmail.com \\
       \addr D-ITET, ETH Z{\"u}rich \\
       Max Planck Institute for Intelligent Systems\\
       \AND
       \name Frederik\ Harder \email frederik.harder@gmail.com \\
       \addr University of T{\"u}bingen \\
       Max Planck Institute for Intelligent Systems\\
        \AND
       \name Mijung\ Park \email mijungp@cs.ubc.ca \\
       \addr Department of Computer Science \\ 
       University of British Columbia\\}

\editor{}

\maketitle

\begin{abstract}
We introduce a simple and intuitive framework that provides quantitative explanations of statistical models through the probabilistic assessment of input feature importance. The core idea comes from utilizing the Dirichlet distribution to define the importance
of input features and learning it via approximate Bayesian inference. The learned importance has probabilistic interpretation and provides the relative significance of each input feature to a model’s output, additionally assessing confidence about its importance quantification. 
As a consequence of using the Dirichlet distribution over the explanations, we can define
a closed-form divergence to gauge the similarity between
learned importance under different models. We use this divergence to study the feature importance explainability tradeoffs with essential notions in modern machine learning, such
as privacy and fairness. Furthermore, BIF can work on two levels: \textit{global} explanation (feature importance across all data instances) and \textit{local} explanation (individual feature importance for each data instance).
We show the effectiveness of our method on a variety of synthetic and real datasets, taking
into account both tabular and image datasets. The code is available at \url{https://github.com/kamadforge/featimp_dp}
\end{abstract}


\clearpage
\section{Introduction}

The increasing spread of machine learning algorithms in a wide range of human domains prompt the public and regulatory bodies to increase scrutiny on the requirements on algorithmic design  \citep{Voigt2017}. 
Among them, \textit{explainability}  aims to provide a human interpretable reasoning for algorithmic decisions \citep{2016arXiv160608813G}. 
With the goal of gaining such explainability, there has been a large number of methods developed for explaining why machine learning models produce particular outputs \citep{Ribeiro2016WhySI, NIPS2017_7062}. In this paper, we tackle the problem of accurately assessing how relevant each of the data input feature is to a machine learning model.

Broadly speaking, there are two levels at which one could gain such input feature explainability. The first level is \emph{global}, where the goal is to identify the most relevant features for a given task globally across all the data instances. However, when the data exhibit a large variability, using global explanation methods is not suitable, as the learned feature importance at the global level is  same for all data samples. 
%
To overcome this limitation, the second level is \emph{local}, where the goal is to identify the most relevant features for each data instance separately \citep{10.5555/3305890.3306006}. 
The current literature proposes separate mechanisms for each of the approaches to feature explanations. The traditional feature selection literature offers a large number of methods for global explanations  \citep{Hall99correlation-basedfeature, 1453511, Candes16, 10.5555/944919.944968, KIRA1992249} while newer methods concentrate on instance-wise approaches \citep{chen2018learning, yoon2018invase}. On the other hand, BIF is a common framework which allows both for global and local explanation.

Providing local explanations is challenging and requires backing by an external model. The current approaches provide two ways for local explanations. The first group provides model-agnostic methods which rely on the outputs of common machine learning models (e.g. Random Forest or XGBoost). The second group concentrates on leveraging neural networks to build their own models which provide feature explanations. BIF is a novel method that bridges both approaches, additionally providing a Bayesian perspective for feature importance. 

Thus, we introduce a framework called
\emph{Bayesian importance of features (BIF)} 
for gaining the explainability of complex machine learning models both globally and locally with the following benefits:

        

\begin{itemize}

    \item It models feature importance via Dirichlet distribution thus providing probabilistic interpretation and relative weighing of the features.
    
    \item By means of Bayesian formulation, it provides the uncertainty measure of the assessed importance values.
    
    \item The algorithm produces the distribution over the feature importance, which we exploit to quantify the trade-offs between a model's explainability in terms of the feature importance and other notions such as privacy and fairness.

    \item As a unified framework, it can learn a global probability over features via a simple model linear in the number of parameters, or apply neural networks to produce an instance-wise explanation.

    \item The proposed method is a flexible meta-algorithm which can work with any model through which we can backpropagate. 
    
    %
    
\end{itemize}






\begin{figure*}[t]
\centering
\includegraphics[width=1\linewidth]{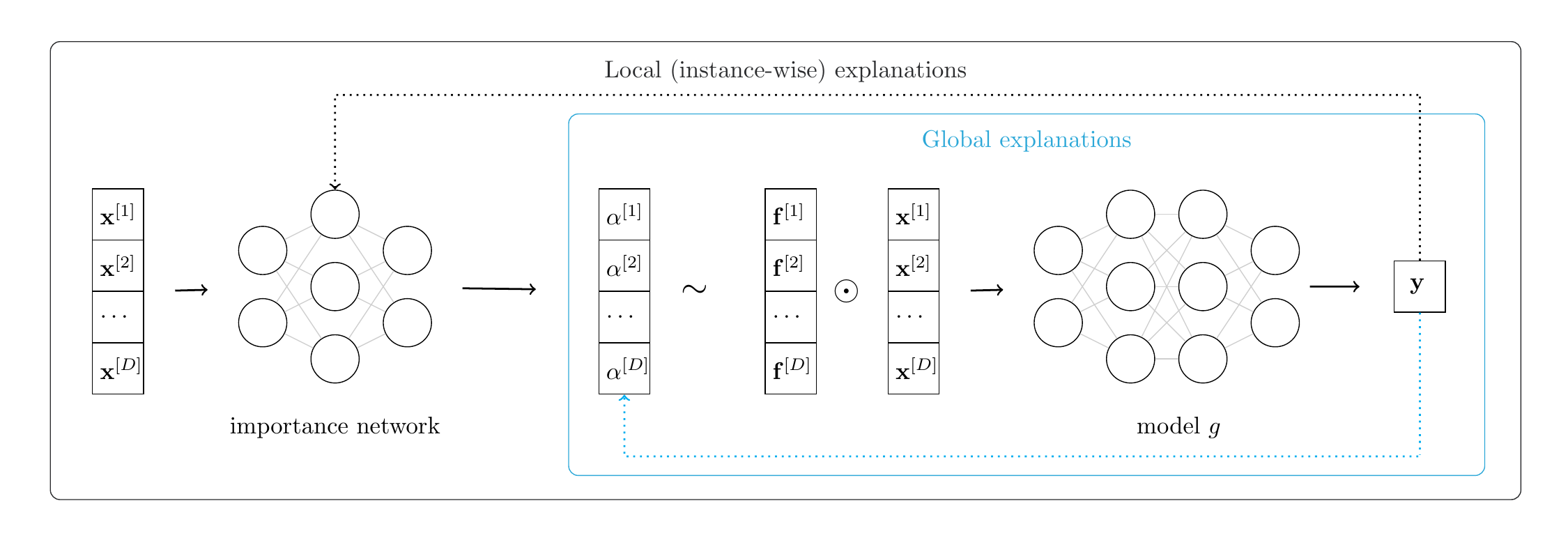}
\vspace{-0.5cm}
\caption{
\textbf{Global explanation} (blue box). We define an importance vector, assumed to be Dirichlet distributed. To learn the parameters of this distribution (denoted by $\alpha^{[1]} \cdots \alpha^{[D]}$), we element-wise multiply a sample of this distribution  ($\vf^{[1]}, \cdots, \vf^{[D]}$) by input features  ($\vx^{[1]}, \cdots, \vx^{[D]}$), which is fed into the model in question (denoted by $g$). The model's prediction $\vy$ and the target determine the loss, used to update the parameters of Dirichlet distribution through  back-propagation (blue dotted line).  \textbf{Local explanation} (black box). To gain instance-wise, local explanation, we define an importance network, where the inputs are the input features ($\vx^{[1]}, \cdots, \vx^{[D])}$) and outputs follow Dirichlet distributions. We learn the parameters of the importance network via back-propagation (black dotted line).
%
Note that the back-propagation does \emph{not} affect the model in question and that we train either the importance vector or the importance network depending on the application.  
}
\label{fig:schematic_qfit}
\vspace{-0cm}
\end{figure*}


\section{Related Work}
\label{sec:Related_Work}

Feature selection, overall, is intended to reduce the number of data input variables to those that are the most useful to a model in order to predict the target variable. The feature selection methods can be subdivided in various ways, supervised and unsupervised methods, or wrapper (which evaluate subsets of variables to maximize performance), filter (which evaluate the relationships between each input feature and the target) and embedded (such as penalized regression or random forest) methods \citep{kuhn2013applied}. Feature importance, the issue which we tackle in this work, belongs to the filter-type methods. Alternatively, we may also distinguish two broad categories, \emph{feature-additive} and \emph{feature-selection} methods \citep{e23010018}. Subsequently, methods can be divided into global explanations which assess the features for all the data, and local (or instance-wise) which provide the explanation for feature in each data instance.

%

%
%
%
Feature-additive methods provide importance of features per dimension, such that their sum matches a quantity of interest, typically the model's output. Two popular such methods are LIME \citep{ribeiro2016should} and SHAP \citep{lundberg2017unified}. 
LIME assumes that any complex model is linear locally. It fits a simple model around a single observation using new samples with permuted features, weighted according to their proximity to the original. SHAP is based on the game-theoretical concept of the Shapley value which assesses average marginal contribution of an input. SHAP, however, utilizes the reinterpretation from \citep{Charnes1988ExtremalPS} where the score is a weighted linear combination of features. This approach provides a local explanation by the level of deviation a given data sample gets from the global feature average. Both LIME and SHAP, similarily to BIF, provide a feature importance weight.
Other notable feature additive approaches include attribution-based methods \citep{abs-1711-06104} such as Integrated Gradient \citep{SundararajanTY17}, Smoothgrad \citep{SmoothGrad}.

On the other hand, feature-selection narrows down the set of features and finds a subset of input features which produce a similar result to the case when the full set of input features is used. Two recent such models are L2X \citep{chen2018learning} and INVASE \citep{yoon2018invase}. One notable difference from the classical feature selection methods such as LASSO \citep{tibshirani96regression} is that these methods target instance-wise feature selection and build their own neural network-based pipeline. L2X maximizes the mutual information between subsets  of features $X_s$ and the response variable $y$, and approximates this quantity with a network which produces binary feature samples learnt via continuous relaxation with the Gumbel-softmax trick. 
INVASE \citep{yoon2018invase} consists of three networks. The first network is a selector network which provides selection probabilities given each input feature.  However, unlike in our model, the outputs of the selector network are treated as separate Bernoulli variables and are not directly trained with the model's output.
The output of the selector network is fed into two separate networks which form an actor-critic pair for feature selection. 
Both of these methods produce binary output of $k$ important features: L2X outputs a predetermined set of important features while INVASE determines the $k$ based on a threshold.

\section{Problem formulation}
\label{sec:problem}

Generally, \textit{feature importance} assigns a score $f \in \mathbb{R}$ to each input feature in the input data $\mathcal{D} \in \mathbb{R}^{N \times D}$, where $D$ denotes the number of features and $N$ the number of data examples. For a given feature, the score can be the same or differ for each data sample $\vx \in \mathcal{D}$, which is described by the two following concepts:

\emph{Global explanation}: We assign a universal, $D$-dimensional vector of values $\textbf{f}$  to the set of input features in the dataset $\mathcal{D}$.
The value quantifies the relevance or importance of the feature for the entire dataset. 
%
In particular, the vector $\mathbf{f}$ assigns a single, scalar-valued global score $\mathbf{f}^{[j]} $ to each feature $j$. 
%

\emph{Local explanation}:  Every instance of data $\vx_n$ is assigned a separate, $D$-dimensional feature importance vector $\mathbf{f}_n$, and the importance vectors $\mathbf{f}_m$ and $\mathbf{f}_n$ for two different data instances $\mathbf{x}_m$ and $\mathbf{x}_n$ need not be the same.

Moreover, feature importance is computed in relation to a model $g$. In other words, a feature is important for a task indicated by a model in question, denoted by $g$, and thus we require a data-model tuple ($\mathcal{D}, g$). 
What comes next describes our method that provides probabilistic interpretation of both global and local explanations.





\section{Methods: Bayesian importance of features (BIF)} 
\label{sec:Methods}

Consider a data-model tuple ($\mathcal{D}, g$) such that $\mathcal{D} \in \mathbb{R}^{N \times D} $ and $g$ is any differentiable model. 
 That is, a model $g$ is trained with an $N$-element dataset $\Dat = \{\vx_n, y_n\}_{n=1}^N$, where $\vx_n \in \mathbb{R}^D$ is an input datum, $y_n$ its label (either discrete or continuous) and $D$ is the input dimension. The proposed method assesses the importance of the set of $D$ input features given the model $g$. Moreover, we aim to idenfity how each feature of the dataset $\mathcal{D}$ affects the output of $g$ on two different levels.

\subsection{Global explanation}
\label{global}

In the global feature explanations, we assign the importance for a feature across the entire dataset. Usually, this value is unbounded \citep{lundberg2017unified} or binary. In our view, it is intuitive to describe the feature importance through a probability vector which describes the relative weight of a feature. Moreover, as we consider the data-model tuple ($\Dat, g$), the importance can be viewed as a contribution of a feature $\mathbf{f}^{[j]}$ to the maximization of the objective function given by the model $g$.
In our method, we name $\vf$ the \emph{importance vector}. This vector is used to perform the scalar product $\mathbf{f} \circ \mathbf{x}$ which becomes the new weighted input to the model $g$. The mechanism is illustrated in \figref{schematic_qfit}.
%

\noindent \emph{\textbf{Loss function for global explanation $\mathcal{L}_G$: }}
Let $\mathcal{L}_G$ denote the loss for obtaining the global feature importance and $p(\mathcal{D})$ the probability of the data under a model $g$. In our method, we assume a parametrized model of joint distribution between the data $\mathcal{D}$\footnote{We preserve the notation $\mathcal{D}$ which is common in the literature, however one should note that in the derivations $\mathcal{D}$ is an equivalent notation for a sample $\vx_n$. For the global case, we omit the index $n$ for clarity.} and the importance feature vector $\vf$. In principle, we want to maximize the log-likelihood of the data which is obtained by integrating out the feature importance vector,
$\log p(\Dat) = \log \int p(\Dat, \vf) d\vf.$
Under a neural network model, directly integrating out $\vf$ is intractable. We instead use an approximate distribution $q(\vf)$, which approximates $p(\vf| \Dat)$. Minimizing the KL-divergence between the two distributions is equivalent to 
maximizing the lower bound to the data log-likelihood:  
%
\begin{align}\label{eq:var_low}
     \LL_G := \int  q(\vf) \log p(\Dat|\vf) d\vf - D_{KL}[q(\vf)|| p(\vf)].
\end{align}
%


\noindent \emph{\textbf{Parameterization}}. The terms in $\LL_G$ are defined as follows:
\begin{align}
q(\vf) &= \mbox{Dir}(\vf|\valpha) \hspace{0.5cm} \text{(approximate posterior)} , \\
p(\vf) &= \mbox{Dir}(\vf|\valpha_0) \hspace{0.5cm} (\text{prior})\label{eq:dir_param}
\end{align}

Both $\valpha$ and $\valpha_0$ are parameter vectors of the Dirichlet distribution. We set the parameters $\valpha_0$ to some constant value and only optimize for $\valpha$. Assuming the Dirichlet distribution both for the posterior and the prior allows us to obtain a closed-form KL-divergence in Eq.~\ref{eq:var_low}. Thus, the objective function in \eqref{var_low} depends on the Dirichlet parameters $\valpha$:
\begin{align}
\label{eq:obj_G_final}
\LL_G(\valpha) := 
    \int  q(\vf|\valpha) \log p(\Dat|\vf) d\vf - D_{KL}[q(\vf|\valpha)|| p(\vf|\valpha_0)].
\end{align}
The crucial characteristic of the global explanation when computing the loss is that for two samples, $\vx_m$ and $\vx_n$, the importance vectors $\vf_m$ and $\vf_n$ are sampled from the same parameters Dir($\valpha$) across the entire dataset. 
%
In the above loss we use the likelihood of the model $g$ (that is $p(\mathcal{D}|\vf)$ which is computed with the output of the model $g$), but we do not alter the parameters of the model $g$ (which we assume to be pre-trained with $\Dat$). We freeze its parameters, and only train the parameters of the feature importance vector $\vf$. The algorithm for obtaining the global feature importance is summarized in \algoref{G}\footnote{Our algorithm is general for any classification (both binary and multi-class) and regression tasks. However, in our experiments we focus on the classification tasks.}.




\noindent \emph{\textbf{Why Dirichlet?}}
Dirichlet distribution describes a family of categorical distributions defined over a simplex, and a sample of the Dirichlet distribution is a probability vector, where all elements of $\vf$ are non-negative and 
$
\sum_{i=1}^D \vf^{[i]} = 1.    
$
This property makes it natural to model a relative level of importance across different input features. Moreover, the choice of Dirichlet distribution allows for the closed-form expression of KL-divergence.


\subsection{Local or instance-wise explanation via Importance Network (IN) \newline}
\label{local}
Local explanations differ from the global ones in that feature importance is evaluated for each data instance. In global explanation, we only produce a single vector $\mathbf{f}$. Conversely, in the local setting, we produce an importance matrix. To be precise, each data point $\vx_n \in \mathcal{D}$ is assigned a vector $\vf_n \in [0,1]^D$  indicating the \emph{feature importance} for that particular data point. Thus, while $|\mathcal{D}|=N$, an importance matrix of size  $N \times D$ is generated.







\noindent \emph{\textbf{Parameterization}}. In the case of local explanations, each of the importance vectors, $\vf_n$ is also modelled by the Dirichlet distribution with individual parameters $\valpha_n$,
\begin{align}
q(\vf_n) &= \mbox{Dir}(\vf_n|\valpha_n) \hspace{0.5cm} \text{(approximate posterior)} \\
p(\vf_n) &= \mbox{Dir}(\vf_n|\valpha_0) \hspace{0.5cm} (\text{prior})\label{eq:local_dir}
\end{align}
where we set the parameters $\valpha_0$ the same for all $\vx_n$, as we do not have any prior knowledge on any particular data instances.

\noindent \emph{\textbf{Importance network.}} The global importance vector $\vf$ tells us the average feature importance  across all the data instances in the entire dataset. However, in the local case, we assign an importance vector $\vf_n$ for each data instance $\vx_n$. To learn the mapping between the two, we resort to an additional model, an \textit{importance network (IN)}\footnote{In our experiments, we use a multi-layer feed-forward network. Note that other types of networks are also possible, e.g. convolutional neural networks can be more appropriate for image data.} parameterized by $\vtheta$. The importance network maps a data instance $\vx_n$ to a corresponding Dirichlet parameter vector $\valpha_n$, i.e., $\mbox{IN}_{\vtheta}: \vx_n \mapsto \valpha_n$. And following Eq. \ref{eq:local_dir}, we draw a corresponding feature importance $\vf_n$ from the Dirichlet distribution with the parameter $\valpha_n$. Hence, the IN model can produce an individual feature importance for each data instance (via Dirichlet parameters). 

Similarly to the global case, we again use the variational lower bound. However, in the local case, our new objective function over $\vf_n$ for all $\vx_n$ become dependent on the parameters of the importance network:
\begin{align}\label{eq:obj_L_final}
\begin{split}
q(\vf) &= \sum_{n=1}^N [ \int  q(\vf_n|\vtheta) \log p(\vx_n|\vf_n) d\vf_n -  \\
 & - D_{KL}[q(\vf_n|\vtheta)|| p(\vf_n|\valpha_0)] ]
\end{split}
\end{align}
%
%
%
%
During training, we set $\valpha_0$ to a fixed value and optimize for $\vtheta$. Given a sample $\vf_n$, we apply an element-wise multiplication,  $\mathbf{f}_n \circ \mathbf{x}_n$, which is fed to the model in question $g$. The model $g$ then produces the conditional distribution $ p(\vx_n|\vf_n) $. 
During learning as in the global case, we do not update the parameters of $g$, but only update the parameters of the IN model. We summarize our algorithm in \algoref{L} and provide the graphical depiction of this process in Fig.~\ref{fig:schematic_qfit} (black box).





%


\begin{figure}[!t]
\begin{algorithm}[H]
\centering
\caption{Global BIF}\label{algo:G}
\begin{algorithmic}[1]
\vspace{0.1cm}
\Require Model in question $\mathit{g}$ with \textit{fixed} weights 
\ForEach {train-mini-batch $b$}
\State Sample $\mathbf{f}$  as in \eqref{dir_param}
\State Compute $g(\mathbf{f} \circ \vx_n)$ for $\vx_n \in b$ 
\State Update $\valpha$ by maximizing $\LL_G$ in \eqref{obj_G_final}.
\EndFor
\State \Return Dirichlet parameters $\valpha$ for global explanation
\end{algorithmic}
\end{algorithm}
\vspace{-0.4cm}
\begin{algorithm}[H]
\centering
\caption{Local BIF}\label{algo:L}
\begin{algorithmic}[1]
\vspace{0.1cm}
\Require Model in question $\mathit{g}$ with \textit{fixed} weights  
\ForEach {train-mini-batch $b$}
\State Compute $\valpha_n$ for each $\vx_n$ using $\mbox{IN}_{\vtheta}$
\State Given $\valpha_n$, sample $\mathbf{f_n}$ from \eqref{local_dir} 
\State Compute $g(\vf_n \circ \vx_n)$ for $\vx_n \in b$
\State Update $\vtheta$ of IN by max $\LL_L$ in \eqref{obj_L_final}.
\EndFor
\State \Return Importance network (IN) parameters $\vtheta$ which outputs local explanations $\valpha_n$ for the input $\vx_n$ in test-mini-batch.
\end{algorithmic}
\end{algorithm}
\vspace{-0.9cm}
\end{figure}

\subsection{Sampling vs.\ a point estimate}
\paragraph{Sampling.} In both objective functions, Eq. \ref{eq:var_low} and Eq. \ref{eq:obj_L_final}, in the left-hand side term (so called, cross-entropy term) we need to evaluate an integral over $\vf$ (or $\vf_n$). We do so by the Monte Carlo integration using Eq. \ref{eq:dir_param} for the global setting and Eq. \ref{eq:local_dir} for the local setting. The integral is evaluated for each data input $\vx_k$ (we use here the $k$ data index to describe both the global and the local case):
\begin{align}
& \int  q(\vf_k|\valpha_k) \log p_g(\Dat|\vf_k) d\vf_k  
\approx  \frac{1}{J} \sum_{j=1}^J \log p(\Dat|\vf_{k,(j)}),
\end{align} 
where the subscript $(j)$ denotes the $j$th Monte Carlo sample $\vf_k$ from the Dirichlet distribution $\valpha_k$. Following \citep{knowles}, we compute the gradients of the integral implicitly using the inverse CDF of the Gamma distribution.
%
%


\paragraph{A point estimate.} A computationally cheap approximation to the integral is using the analytic mean expression of the Dirichlet random variables, 
\begin{align}
     \int q(\vf) \log p(\Dat|\vf) d \vf |_{\vf = \bar{\vf}} &\approx \log p(\Dat|\bar{\vf}),
\end{align} where $\bar{\vf} = \frac{\valpha}{\sum_{d=1}^D\valpha^{[d]}}. $
Computing the point estimate does not require sampling and propagating gradients through the samples, which significantly reduces the run time. In our experiments, we use both approximations where the specifics on each approximation are included in the Supplementary material. 



\subsection{Divergence for measuring similarity under BIF }

The BIF's output, the Dirichlet distribution is an exponential family distribution, which we can write in terms of an inner product between the sufficient statistic $T(\vf)$ and the natural parameter $\veta$:  
\begin{align}
p(\vf|\valpha) &= h(\vf) \exp \left[ \langle \veta(\valpha), T(\vf) \rangle - A(\veta) \right]
\end{align} where $A(\veta) = \log \int h(\vf) \exp( \langle \veta(\valpha), T(\vf) \rangle ) d\vf $ is the log-partition function, and $h(\vf)$ is the base measure.
In case of the Dirichlet distribution, the natural parameter equals the parameter  $\veta(\valpha) = \valpha$, yielding a canonical form. 

We are interested in measuring how similar feature importance is under two different models. We denote the two Dirichlet distributions for feature importance obtained under the two models by $p$ and $q$, respectively, where $p$'s parameters are $\bm{\alpha} = [\alpha^{[1]}, \cdots, \alpha^{[D]}]$ and $q$'s are $\bm{\beta} = [\beta^{[1]}, \cdots, \beta^{[D]}]$. 
Luckily under the exponential family distribution, popular divergence definitions such as the KL divergence $D_{KL}(p||q)$ and the Bregman divergence $B(q||p)$ can be  expressed in terms of the log-partition function, its parameter, and the expected sufficient statistic:
\begin{align}
    &D_{KL}(p||q) = B(q||p) \nonumber \\
    &:=  A(\veta(\valpha)) - A(\veta(\vbeta)) - \langle \valpha-\vbeta, \mathbb{E}_p[T(\vf)] \rangle, 
\end{align} 
where $\mathbb{E}_p[T(\vf)]$ is the expected sufficient statistic under the distribution $p$.
Under the Dirichlet distribution, all of the three terms are in closed-form, where the log-partition function is defined by 
$A(\veta(\valpha)) = \log \Gamma(\alpha_0) - \sum_{d=1}^D \log \Gamma(\alpha^{[d]}), 
A(\veta(\vbeta)) = \log \Gamma(\beta_0) -  \sum_{d=1}^D \log \Gamma(\beta^{[d]})$, where $\Gamma$ denotes Gamma distribution,
and each coordinate of the expected sufficient statistic is defined by 
$\mathbb{E}_p[T(f_d)] = \psi(\alpha^{[d]}) - \psi(\sum^D_{d} \alpha^{[d]})$, where $\psi$ is the digamma function. This allows us to  evaluate the KL divergence conveniently. 
We demonstrate how we take advantage of having this easy-to-evaluate divergence in practice and examine the results under BIF in \secref{Experiments}.

\section{Experiments}
\label{sec:Experiments}

\input{Experiments}
\section{Conclusion} \label{sec:conclusion}
Our novel Bayesian perspective yielded the framework that is accurate and also provides confidence about the feature importance both in global and local settings. The learnt distributions over features can be useful to measure explainability in a variety of applications.



\newpage

\appendix

\label{app:theorem}

\section{Experiment details for synthetic data}

Below we give details on the experimental setups for each of the results presented in the paper.

\subsection{Methods}

\paragraph{BIF.} We first train the network for 500 epochs. Then we freeze these weights and finetune the switch vector or switch network for 10 epochs. In the experiment 1 we use analytic mean of Dirichlet distribution. We also tested the sampling but analytic mean proved to work faster and better.

\paragraph{L2X.} We load the additional datasets for 125 epochs (preserving number iterations due to the smaller dataset). On Syn4 and Syn5 where the number of relevant features is not fixed, we report results for k=5, which maximized $(TPR-FDR)$.

\paragraph{INVASE.} We run the original code for 10k iterations. As suggested by the authors (\url{https://github.com/jsyoon0823/INVASE/issues/1}), we set $\lambda=0.15$ for Syn5 and to $0.1$ otherwise. We use SELU nonlinearities for Syn4-6 but not for Syn3, as we obtained better results with ReLUs on the first three datasets. 

\subsection{Standard deviation comparison.}

We compare here the standard deviation of the selected methods used in the main text.

\begin{table}[h]
\centering
\begin{tabular}{ccccccc}
\toprule
& Syn 1 & Syn 2 & Syn 3 & Syn 4 & Syn 5 & Syn 6 \\
\midrule
BIF (inst, samp) & 0 & 0 & 0.107 & 0.019 & 0.010 & 0.025 \\
BIF (inst, pe) & 0 & 0 & 0.201 & 0.015 & 0.061 & 0.093 \\
BIF (gl, samp) & 0 & 0 & 0.201 & NA & NA & NA \\
\bottomrule
\end{tabular}
\caption{The comparison of the standard deviation of the MCC (Matthews correlation coefficient) (in 5 runs) presented in the Table 1 in the main text. For the global case we do not observe variability due to very few parameters and consistent convergence of the algorithm}
\end{table}

\begin{table}
\scalebox{0.9}{
\begin{tabular}{cccccc}
\toprule
& $k=1$ & $k=2$ & $k=3$  & $k=4$ & $k=5$ \\
\midrule
\textbf{BIF} & $\mathbf{0.788 \pm 0.038}$ & $\mathbf{0.937 \pm 0.039}$ & $\mathbf{0.973 \pm 0.008}$ & $\mathbf{0.98 \pm 0.005}$ & $\mathbf{0.981 \pm 0.004}$ \\
\textbf{L2X} & $0.633 \pm 0.109$ & $0.761 \pm 0.059$ & $0.84 \pm 0.079$ & $0.871 \pm 0.066$ & $0.864 \pm 0.049$ \\
\textbf{INVASE} & $0.584 \pm 0.096$ & $0.78 \pm 0.037$ & $0.901 \pm 0.013$ & $0.915 \pm 0.004$ & $0.905 \pm 0.029$ \\
\bottomrule
\end{tabular}
}
\caption{Quantitative performance of BIF and the corresponding benchmarks on the MNIST image dataset. Post-hoc accuracy of MNIST classifier distinguishing digits 3 and 8 based on $k$ number of (4x4) selected patches. BIF outperforms other methods. This is a detailed version of 
Figure 2
where $\pm$ denotes one standard deviation. This was omitted in the main text due to space limitations.
}
\end{table}
\section{Experiment details for real-world data}

\subsection{Tabular data information}

\paragraph{BIF.} We perform experiments on three real-world tabular datasets, \textit{adult},  \textit{credit} and \textit{intrusion}. We included more details about these datasets and experimental details below. In these experiments we use the point estimate without KL-regularizer.

\subsubsection*{Credit}

Credit card fraud detection dataset contains the categorized information of credit card transactions which were either fraudelent or not. Ten dataset comes from a Kaggle competition and is available at the source,
\url{https://www.kaggle.com/mlg-ulb/creditcardfraud}. The original data has 284807 examples, of which negative samples are 284315 and positive 492. The dataset has 31 categories, 30 numerical features and a binary label. We used all but the first feature (Time).

\subsubsection*{Adult}

The dataset contains information about people's attributes and their respective income which has been thresholded and binarized. It has 22561 examples, and 14 features and a binary label. The dataset can be downloaded by means of SDGym package,\url{https://pypi.org/project/sdgym/}. 

\subsubsection*{Intrusion}

The dataset was used for The Third International Knowledge Discovery and Data Mining Tools Competition held at the Conference on Knowledge Discovery and Data Mining, 1999, and can be found at \url{http://kdd.ics.uci.edu/databases/kddcup99/kddcup99.html}. We used the file, kddcup.data\_10\_percent.gz. It is a multi-class dataset with five labels describing different types of connection intrusions. The labels were first grouped into five categories and due to few examples, we restricted the data to the top four categories. 

\subsection{Standard deviation information for Table 2}

\begin{table*}[h!]
\centering
\scalebox{0.9}{
\begin{tabular}
{c | ccc | ccc | ccc }
\toprule
 & & Adult & & & Credit & & & Intrusion & \\
$k$ & 1 & 3 & 5 & 1 & 3 & 5 & 1 & 3 & 5  \\ \midrule



\textbf{BIF (Local)} 
& 1.547 & 0.581 & 0.655 
& 3.804 & 1.217 & 2.737 
& 2.269 & 8.874 & 3.926 \\

\textbf{BIF (Global)} 
& 1.220 & 5.88 & 0.000
& 1.579 & 0.839 & 0.200
& 4.712 & 6.133 & 3.128  \\
 




\bottomrule
\end{tabular}}
\vspace{0.3cm}
\caption{\textbf{Tabular datasets}. Classification accuracy standard deviation as a function of $k$ selected features. Complementary information to the Table 2 in the main paper. The variability in the results are presented for the local case where we train the importance network. In the global case, the ranking of parameters which is equal to the number of features is converging and is consistent and therefore we do not observe the variability.}
\label{tab:real}
\vspace{-0.15cm}
\end{table*}

\subsection{MNIST data}

We generate the binary classification dataset by selecting all samples of classes 3 and 8 from MNIST, keeping the separExperimentracy on the feature-selected data. In all cases below, the feature selection models output 49 dimensions representing the patches, which are then copied over 4x4 pixel patches to a full output size of 784.

\paragraph{BIF.} We first train a classifier for 10 epochs with the same architecture as the post-hoc accuracy model, but without batch-norm, as this leads to more stable selector training. Following this, the selector is network is trained for 10 epochs and then we use the selector to generate feature importances for the full test-set. For each sample the $k$ most highly weighted patches are kept and the remaining features are set to 0. 

\paragraph{L2X.} Because the original released L2X code does not contain the setup for the MNIST experiment, we use our own implementation based on the released code for synthetic data. We increase hidden dimensions in the selector and classifier parts of the model from $100$ \& $200$ to $250$ \& $500$ due to the higher data complexity.  

\paragraph{INVASE.} We adapt the INVASE setup for synthetic data to mnist and tune the $\lambda$ parameter in order to produce feature selections with different average numbers $k$ of selected features. For $k$ ranging from 1 to 5, we use $\lambda$ values of 100, 50, 23, 18.5, 15.5. As the relationship between the value of $\lambda$ and average $k$ is not reliable, we discard results from random seeds that didn't produce the desired $k$. The chosen random seeds are listed in the experiment code.

\section{Importance value estimates short analysis}

In the paper, we mention that BIF's importance values are more accurate due than those from the existing literature, that is we look here not whether an input feature is important but how important it is. Here we briefly elaborate on this statement and provide some examples. This sort of analysis can be done on synthetic datasets (Syn 1-3) where we know the ground truth and know how the features were generates. Thus, Syn1 and Syn2 are the datasets where the weight of each of the important feature is equal and so we would expect each relevant weight to be equal. In fact, this is what we see in the output of BIF. The global weights for Syn1 are $[0.4986, 0.5015]$ and for Syn2 are $[ 0.2585, 0.2490,0.2451,0.2449]$, both of which very closely discover the weight importance. Furthermore, we provide below a comparison example with the other methods for a more challenging Syn3.

\begin{figure}[h]
\centering
\begin{minipage}[l]{0.4\textwidth}
\includegraphics[scale=0.3]{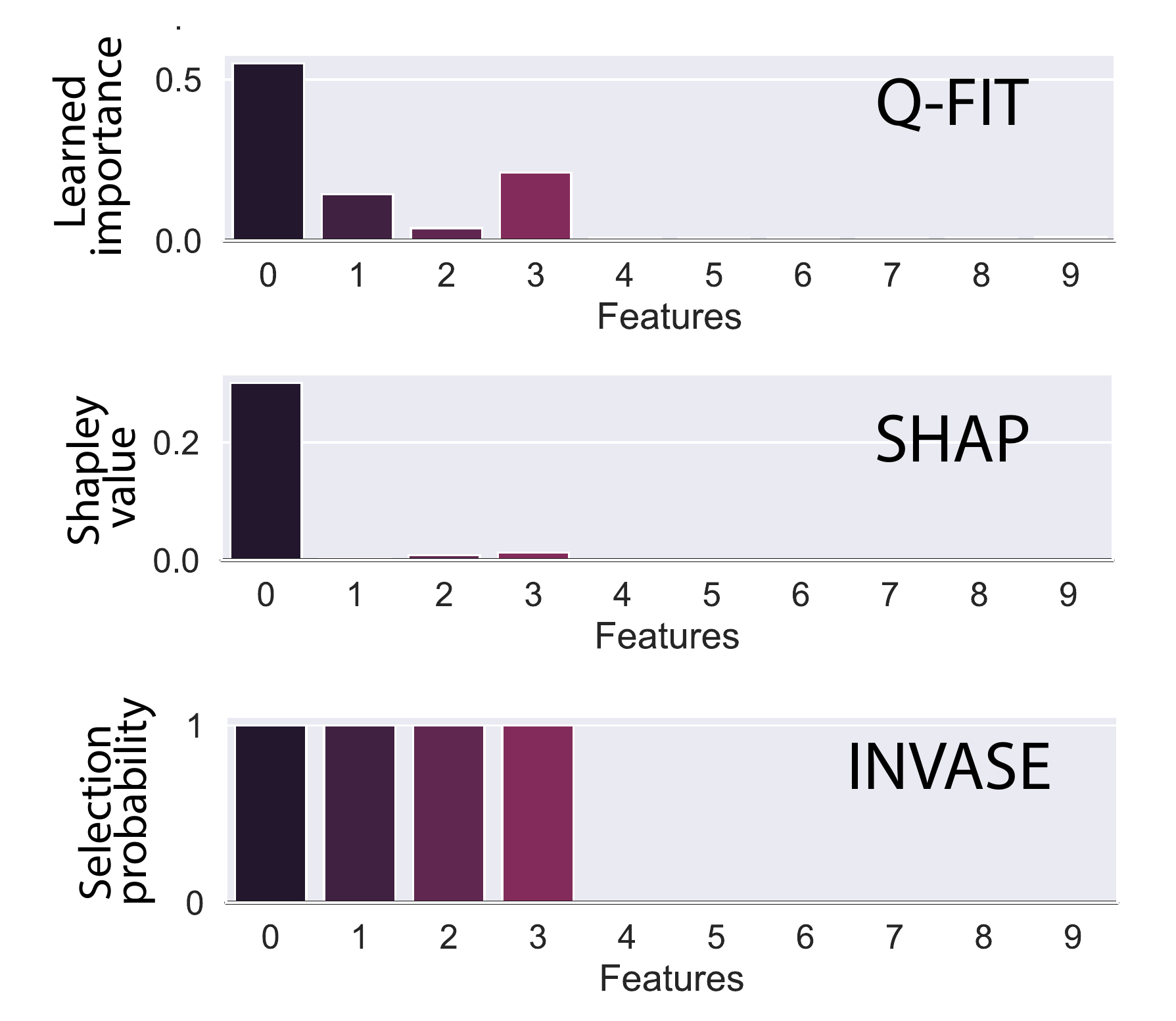}
\end{minipage}
\begin{minipage}[r]{0.4\textwidth}
\caption{Learned feature importance using a dataset with input features $\vx \sim \Nrm(0,I)$ where $\vx \in \mathbb{R}^{10}$ and $p(y=1|\vx) \propto
\exp[-100 \sin({2 X_0})+2|X_1|+X_2+\exp(-X_{3})]$, following \citep{chen2018learning}. 
 \textbf{Top}: Our method uncovers the ground truth correctly with a different level of importance for the four features. 
 \textbf{Middle}: SHAP performed similarly as ours.
 \textbf{Bottom}: INVASE's selection probability (the probability of Bernoulli random variables) is all equal for selected features, giving less information about the feature importance than our method.
 } \label{fig:global}
\end{minipage}
\end{figure}

\section{Privacy trade-off analysis for INVASE}

\begin{figure*}[h!]
\includegraphics[width=\textwidth]{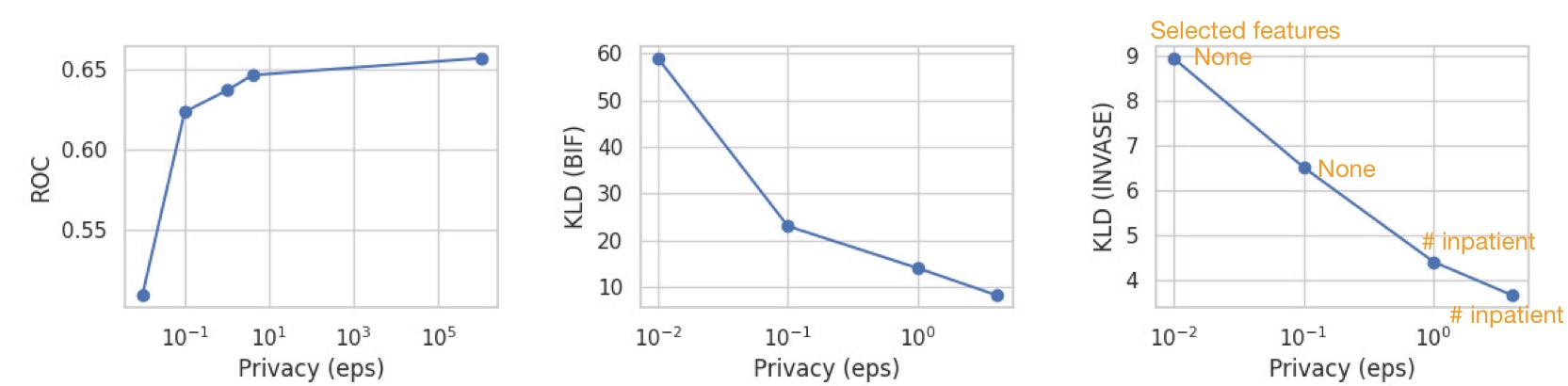}
\caption{\textbf{Privacy vs feature importance on Diabetes Readmission data}\citep{rizvi}.
\textbf{Left:} Trade-off between accuracy (in terms of ROC) and privacy. The stronger the privacy level, the worse the classifier's performance. \textbf{Middle:} The feature importance learned by BIF at a stronger privacy (small $\epsilon$) has a larger divergence from the feature importance learned non-privately. %
\textbf{Right:} While KL divergence shows a similar pattern as the middle plot, the difference in the KL divergence under INVASE lacks consistency with the learned importance of features (selected features by INVASE are written in Organge). For instance, at $\epsilon=0.01$ and $\epsilon=0.1$, INVASE selected no features as important. On the other hand, at $\epsilon=1.0$ and $\epsilon=4.0$  INVASE selected the number of inpatient as an important feature.
} 
\vspace{+0.2cm}
\label{fig:privacy}
\end{figure*}

For INVASE  \citep{yoon2018invase}, we adjust the setup so that we jointly train the baseline network with a private classifier, then freeze the baseline network, and only update the selector and predictor networks.
As INVASE is a method for instance-wise feature selection, once trained, we use INVASE to output the feature selection for the test datapoints, and average the selection probability  (the Bernoulli distribution over the feature selection)  across those test datapoints. 
%

The \textbf{Middle} and \textbf{Right} plots in \figref{privacy} compare INVASE and BIF. In the case of INVASE, the KL divergence is not necessarily informative. The divergence metric between the selection distribution under the non-private classifier and that at different privacy level differ, while the selected features remains similar. 

\textbf{Right:} While KL divergence shows a similar pattern as the middle plot, the difference in the KL divergence under INVASE lacks consistency with the learned importance of features (selected features by INVASE are written in Orange). For instance, at $\epsilon=0.01$ and $\epsilon=0.1$, INVASE selected no features as important. On the other hand, at $\epsilon=1.0$ and $\epsilon=4.0$  INVASE selected the number of inpatient as an important feature.

\section{Fainess trade-off analysis}

\begin{figure*}[h!]
\includegraphics[scale=0.2]{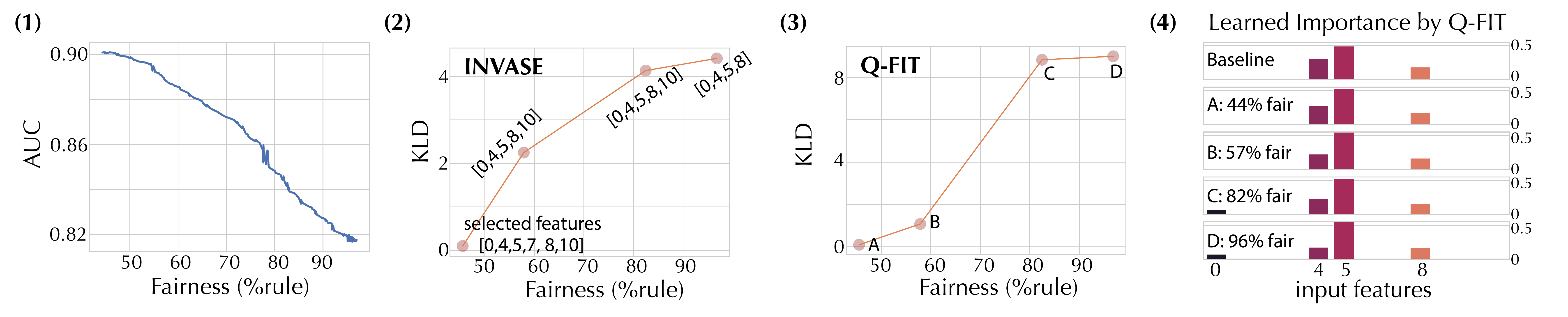}
\caption{\textbf{Fairness vs feature importance on modified Adult data.} \textbf{(1)} Trade-off between accuracy and fairness of a classifier. 
\textbf{(2)} We evaluate the KL divergence of selection probabilities between  the baseline (unfair) classifier and varying levels of fair classifiers ($44, 57, 82$ and $96\%$).
The 2nd and 3rd dots show the selected features by INVASE being identical while their KL divergence values differ.
\textbf{(3)} The KL divergence between the BIF's importance distributions at varying levels of fair classifiers and the baseline classifier.  \textbf{(4)} The difference in KL divergence under BIF is well reflected in the learned importance at different levels of fairness. } 
\label{fig:fairness}
\end{figure*}

\textbf{Fairness vs feature importance. } 
First, we show the usefulness of our method to study the trade-off between explainability in terms of feature importance and fairness. 
We consider a fair classifier introduced in \citep{url_fair_clf}, and the Adult data to train a fair classifier in terms of Race. For this experiment, we modify the dataset, such that it only consists of $12$ features by excluding the Race and Sex features as done in \citep{url_fair_clf}. 
As shown in \textbf{(1)} in \figref{fairness}, the classifier loses accuracy measured in terms of the area under the curve as we increase the fairness measured in terms of the percentage rule \citep{url_fair_clf}, which is a well-known phenomenon, but we aim to show that BIF allows to measure well the loss in explainability when increasing the fairness of the classifier.

%

In this case, as shown in \textbf{(2)} and \textbf{(3)} of \figref{fairness}, both INVASE and BIF demonstrate gradual loss of explainability as the KL divergence between the feature distribution under the classifier trained without any fairness constraint and that at different levels of fairness ($44\%, 57\%, 82\%$ and $96\%$ fair) increases as the fairness constraint increases. However, the vanilla INVASE by outputting just a set of important features does not distinguish between the level of $57\%$ and $82\%$ (by outputting $[0,4,5,8,10]$ for both levels). Meanwhile, under BIF, the difference in the KL divergence is well reflected in the learned importance as shown in \textbf{(4)}. By virtue of assigning continuous importance weights, BIF is able to account for smaller changes in the trained model than a discrete method like INVASE.

\section{Experiment details for Trade-offs}

\subsection{Feature importance vs fairness}

\paragraph{Dataset} We use the Adult data used in the earlier section, with a change of excluding sex and race from the input features. The resulting dataset has $12$ input features: age(0), workclass(1), fnlwgt(2), education(3), education number (4),  marital status(5), occupation(6), relationship(7), capital gain(8), capital loss(9), hours per week(10), native country(11). 

\paragraph{Fair classifiers} 
We use this modified dataset and train a classifier with different fairness constraints following code from \url{https://github.com/equialgo/fairness-in-ml/blob/master/fairness-in-ml.ipynb }

\paragraph{BIF} Given a classifier at each level of fairness, $44,57,82,96\%$ in terms of percentage rule based on race, we learn the global feature importance vector for $400$ epochs. When computing the variational lower bound, we used a single sample.  We set the prior parameter to be $\alpha_0 = 0.1$.  As shown in the main text, initially when the classifier is unfair, the marital status(5) is the most important, and  the education number (4) is in the second place, and the capital gain (8) is in the third place. As we increase the fairness in training the classifier, while the importance of marital status remains almost the same, the education number and capital gain become less important, while the age starts appearing to be important.   

\paragraph{INVASE} We train the INVASE model by exchanging the classifier component with the pre-trained fair models and only optimize the selector component. This diverges from the standard way of training INVASE, but despite the frozen classifier the model achieves an accuracy of 83\% by the end of training.

\subsection{Feature importance vs privacy}

\paragraph{Dataset} We use the original Adult dataset, which contains $14$ input features: age(0), workclass(1), fnlwgt(2), education(3), education number (4),  marital status(5), occupation(6), relationship(7), race(8), sex(9), capital gain(10), capital loss(11), hours per week(12), native country(13).

\paragraph{Private classifiers}

Using this dataset, we train a classifier, a 3-layer feedforward network with $100$ and $20$ hidden units in each hidden layer for $20$ epochs, using the differentially private stochastic gradient descent (DP-SGD), which adds appropriately adjusted amount of noise to the gradient during training for privacy. The amount of noise and the corresponding privacy guarantee of the classifier is summarized in Table \ref{tab:privacy_noise}. The highly nonlinear relationship between the noise level and the corresponding privacy level is calculated by using the autodp package:  \url{https://github.com/yuxiangw/autodp}. 

\begin{table}[!h]
    \centering
     \begin{tabular}{cc}
    \midrule
        noise level & privacy guarantee \\
        \hline
        $\sigma = 0$ & $\epsilon = \infty$\\
        $\sigma = 1.35$  & $\epsilon = 8.07$  \\
        
       $\sigma = 2.3$  & $\epsilon = 4.01$  \\
        $\sigma = 4.4$  & $\epsilon = 1.94$  \\
         $\sigma = 8.4$  & $\epsilon = 0.984$  \\
          $\sigma = 17$  & $\epsilon = 0.48$  \\
              
          \bottomrule
    
    \end{tabular}
    \caption{Differential privacy guarantees based on noise levels  
    }
    \label{tab:privacy_noise}
\vspace{-0.4cm}
\end{table}



\paragraph{BIF} Under each of the classifier, we learn the global feature importance of the input features using BIF for $400$ epochs. When computing the variational lower bound, we used a single sample. We set the prior parameter to be $\alpha_0 = 0.01$. Under the non-private classifier, relationship(7) status is the most important, and education number (4) is  second, and capital gain(10) is the third. As we increase the noise level for a stronger privacy guarantee, the importance of the relationship(7) feature gets lower and other features such as age(0) and marital status(5) become more important. 

\paragraph{INVASE} The setup here equals the setup for fair models, with the difference that privacy classifier models are loaded instead.

\section{Matthews correlation coefficient}

Matthews correlation coefficient that is used in the Experiments section of the main text is defined as follows: 

$$\text{MCC}={\frac {{\mathit {TP}}\times {\mathit {TN}}-{\mathit {FP}}\times {\mathit {FN}}}{\sqrt {({\mathit {TP}}+{\mathit {FP}})({\mathit {TP}}+{\mathit {FN}})({\mathit {TN}}+{\mathit {FP}})({\mathit {TN}}+{\mathit {FN}})}}}$$

\section{Hardware}

We have implemented our experiments in PyTorch \citep{pytorch} on a laptop with GeForce RTX 2080. In our experiments we also used Nvidia Kepler20 and Kepler80 GPUs, or a cluster consisting of five Tesla K80 and GeForce RTX 2080 Ti. 
\clearpage

\vskip 0.2in
\bibliography{bibliography}

\end{document}

%% file: notation.tex
\newcommand{\punt}[1]{}

\usepackage{amsthm}
\usepackage{xcolor}
\usepackage{hyperref}

\renewcommand{\eqref}[1]{Eq.~\ref{eq:#1}}
\newcommand{\Nrm}{\mathcal{N}}

\newcommand{\figref}[1]{Fig.~\ref{fig:#1}}  
\newcommand{\secref}[1]{Sec.~\ref{sec:#1}}
\newcommand{\tabref}[1]{Table ~\ref{tab:#1}}  
\newcommand{\algoref}[1]{Algorithm~\ref{algo:#1}}  

\newcommand{\vx}{\mathbf{x}}

\newcommand{\Dat}{\mathcal{D}}

\newcommand{\vf}{\mathbf{f}}

\newcommand{\valpha}{\mathbf{\ensuremath{\bm{\alpha}}}}
\newcommand{\vbeta}{\mathbf{\ensuremath{\bm{\beta}}}}
\newcommand{\veta}{\mathbf{\ensuremath{\bm{\eta}}}}

\newcommand{\vtheta}{\mathbf{\ensuremath{\bm{\theta}}}}
\newcommand{\LL}{\ensuremath{\mathcal{L}}}

\newcommand{\vy}{\mathbf{y}}

%% file: Experiments.tex
\input{table11}
\input{table1}

We perform the experiments on both synthetic and real-world datasets. The binary synthetic datasets are meant to show the accuracy in selecting the appropriate features which were used to impact the label. The real-world datasets consist of both binary and multi-class datasets, including tabular and image data, and are meant to show broad applicability of the method. Finally, we present the need for well-tuned feature importance probabilities in privacy vs. explainability trade-off. In the experiments, we use the state-of-the-art benchmarks for comparison which allow for the instance-wise feature selection, that is L2X \citep{chen2018learning}, INVASE \citep{yoon2018invase}, SHAP \citep{lundberg2017unified} and LIME \citep{ribeiro2016should}.
%

\subsection{Synthetic data}
\label{sec:synthetic_data}

We first test our method on six synthetic datasets with the aim to identify the relevant features. We construct a data vector $\vx$ in such a way that it is a random variable vector, $\bm{X}=[X^{[1]}, X^{[2]}, \dots X^{[D]}]$, where the index describes an input feature. The first three binary synthetic datasets \citep{chen2018learning} contain a fixed set of relevant features to test the global feature selection.
Each data point consists of a 10-dimensional input feature $\bm{X} \sim \mathcal{N}\mathbf{(0,I)}$ and the associated label that depends on a subset of its features in such a way that $p(y=1|\bm{X})=\frac{1}{1+r}$ and $p(y=0|\bm{X})=\frac{r}{1+r}$ where
the particular $r$ is defined by 
\begin{itemize}[noitemsep]
\item \textbf{Syn1: } $\exp(X^{[1]}X^{[2]})$, 
\item  \textbf{Syn2: } $\exp(\sum_{3}^{6} (X^{[i]})^2-4)$, 
\item \textbf{Syn3: } $\exp(-100 \, \sin{(2X^{[7]})}+2 |X^{[8]}|+X^{[9]}+\exp(-X^{[10]}))$.
\end{itemize}
In the remaining three datasets, we introduce an extra variable, $X^{[11]}$ which selects which set of features determine the label $y$, thus indirectly influencing the result, as well. As a result, a label $y$ depends on an alternating set of features which tests for local feature selection where a set of relevant features varies across a dataset:
%
\begin{itemize}[noitemsep]
\item \textbf{Syn4:} if $X^{[11]}<0$, sampled from Syn1, else Syn2.
\item \textbf{Syn5:} if $X^{[11]}<0$, sample from Syn1, else  Syn3.
\item \textbf{Syn6:} if $X^{[11]}<0$, sampled from Syn2, else Syn3.
\end{itemize}

Notice that the features of the two datasets do not overlap and we can uniquely distinguish the features which generated a given sample.  We generate 10,000 samples for each dataset, using $80\%$ for training and the rest for testing. 
%
Note that in case of Syn1-3, these features are static (suitable for studying global feature importance), while in the case of Syn4-6, the features are alternating (suitable for studying instance-wise feature importance). As we aim to identify the relevant features, we use the \textit{Matthews correlation coefficient (MCC)} \citep{matthews75}. 

\tabref{table_synthetic} summarizes how each of the algorithms copes to uncover the ground truth important features in the synthetic datasets. We include three variations of the proposed method, global selection described in Sec.\ \ref{global} and two ways to compute local explanations described in Sec.\ \ref{local}. The first one evaluates the full integral through sampling (which we denote by samp), while the second one uses a point estimate (which we denote by pe) computed analytically as a mean of the Dirichlet distribution parameters. In all of these variants, we first pre-train a model $g$, and then feed it to our framework where we freeze the model parameters, and only optimize the importance parameters. The proposed method excels particularly in the local setting, on a more challenging datasets, where in all three datasets outperforms the existing methods by a substantial margin, 10-20 percentage points. 

\paragraph{Uncertainty.} In the experiment featuring synthetic datasets, we also verify how well BIF can estimate the uncertainty of the importance values. While non-probabilitic methods provide only the point estimate, Bayesian approach allows to estimate how confident the algorithm is about its output, in this case, the feature importance value. Thus,
Fig. 2 shows the posterior mean and variance of the importance vector. The posterior variance stands for the \textit{confidence} the BIF algorithm has about the learned feature importance. Fig. 2 shows the variance of two datasets, Syn2 which is easier and Syn3 which is harder to predict its label. BIF consequently indicates the higher uncertainty for the harder dataset and lower for the easier dataset. This information can be particularly helpful when assessing the confidence about the importance of each feature, for example, it answers the question how probable it is that a given feature is the most important.

 \paragraph{Discussion.} In selecting feature selection method, it is worth considering their advantages and disadvantages. The BIF global is a method whose number of parameters is linear in the number of features, however it works only in the global setting. The local variants require an additional network which may however work better, also in the global setting (by averaging the importance over all the data points). In terms of performance, the sampling and point estimate BIF produce similar results, however the shortcoming of the point estimate is that it produces the results with relatively high variance (see Supplementary materials for summary results). On the other hand, sampling is more time-consuming due to evaluations required for each sample. It is also worth discussing the effect of features on the label in synthetic datasets. In particular, Syn1 and Syn2 datasets affect the label positively, while in Syn3 we deal with a sine term which may affect the label both positively and negatively. Despite that, the feature has been detected successfully to large extent. Nonetheless, Syn3 is the most challenging dataset among the three synthetic global datasets.



\input{table2}

\subsection{Real-world data}
\label{sec:realworld}

%
\paragraph{Tabular Data} We consider \textit{credit} \citep{Credit} (license: DbCL v1.0) and \textit{adult} \citep{Dua:2019} datasets with tabular input features and binary labels, and \textit{intrusion} \citep{Intrusion} dataset with multi-class labels. Adult dataset predicts whether income exceeds \$50K a year based on census data, credit dataset classifies applicants for credit availability, and the intrusion dataset classifies several types of burglaries. 

%
%
This experiment consists of two parts. In the first part, we look for globally important  features in the entire dataset. In the second part,we perform local feature search. 
%
As there is no known ground truth about the features, we evaluate the effectiveness of each method by selecting top $k$ features which are deemed most significant, and then performing the post-hoc classification task given these $k$ input features with removing the rest. In the global setting, $k$ features are fixed for the entire dataset, while in the local setting each sample can select a different set of $k$ features. In the experiments, we standardize each feature to have a zero mean to mitigate the issue of out-of-distribution examples which could occur in the case of non-zero mean real-world features.

BIF outputs a probability distribution which directly allows to identify top $k$ features. On the other hand, INVASE and L2X output binary decisions for feature importance. For global explanations, we average the output for all data points, thus creating global ranking of features in INVASE and L2X. For local explanations, in case of L2X we can specify $k$ relevant features. INVASE has no such option and thus, for the fairest comparison, we use the selection probability given by the selector network as a proxy of importance score. In the global case of LIME, we average the rankings for the individual instances.
%
As shown in \tabref{real} (classification accuracy averaged over five independent runs), BIF performs well in both tasks, with a bigger edge in global search. 
We found that local explanation search is significantly more challenging than the global one, reflected in the lower classification accuracy. And so although the results provide good insight into which features are important locally, one should proceed with caution when relying on a subset of local features, especially in more risk-averse applications. 



Please also note that as we use a separate validation set to test the top $k$ features, increasing the number of features may have a contrary effect and actually decrease the test accuracy. This happens irrelevant of the method,  likely due to correlation between features and the nature of the model $g$.

\begin{figure}
  \begin{minipage}[b]{0.60\linewidth}
   
        \centering
        \small
\begin{tabular}{cccccc}
\toprule
& $k=1$ & $k=2$ & $k=3$  & $k=4$ & $k=5$ \\
\midrule
\textbf{BIF} & \textbf{0.788} & \textbf{0.937} & \textbf{0.973} & \textbf{0.98} & \textbf{0.981} \\
\textbf{L2X} & 0.633 & 0.761 & 0.84 & 0.871 & 0.864 \\
\textbf{INVASE} & 0.584 & 0.78 & 0.901 & 0.915 & 0.905 \\
\bottomrule
\end{tabular}
  \end{minipage}%
  \begin{minipage}{0.35\linewidth}
    \includegraphics[width=1.2\textwidth]{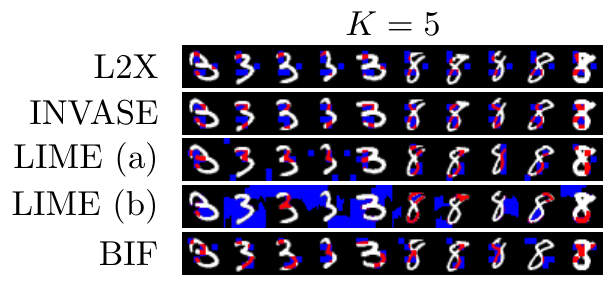}
\end{minipage}
\caption{Quantitative and qualitative performance of BIF and the corresponding benchmarks on the MNIST image dataset. (Left) Post-hoc accuracy of MNIST classifier distinguishing digits 3 and 8 based on $k$ number of (4x4) selected patches. BIF outperforms other methods. 
(Right) Qualitative comparison of feature selection methods for $10$ randomly selected instances of MNIST digits. The selected patches are highlighted in red and blue color.
We show results for LIME both with our pre-determined segmentation into 4x4 patches (a) and using its own segmentation of the image pixels (b). 
BIF seems to particularly well include the differentiating curves between the digits 3 and 8 (see first, third, and fifth digit 8).
}
\label{fig:mnist}
\vspace{-0.3cm}
\end{figure}

\paragraph{MNIST Data.}
Following \citep{chen2018learning}, we construct a dataset with two labels by gathering the 3 and 8 digit samples from MNIST \citep{lecun2010mnist} (license: CC BY-SA 3.0). We then train BIF, as well as L2X and INVASE models to select 4x4 pixel patches as relevant features. As the inputs have a dimensionality of 28x28, there are 49 features to choose from. To evaluate the quality of the selection, we first mask the test set by setting all non-selected patches to 0 and then use a classifier which was trained on unmasked data to compute the \textit{post-hoc} accuracy on this modified test set. The post-hoc accuracies averaged over 5 runs are shown in \figref{mnist}a for different numbers of $k$ selected features.

The selection method differs between models. For BIF, we select the $k$ most highly weighted patches and for L2X, $k$ is set in advance. INVASE is treated differently, as the number of selected features varies and can only be modified implicitly through the strength of the regularizer. So we tune the regularizer strength $\lambda$ to different values such that the average number of selected features equals $k$. The $\lambda$ values we use are 100, 50, 23, 18.5 and 15.5 for $k=1,...,5$.
We also show the qualitative results of each method in  \figref{mnist}b.

\subsection{Divergence for comparing feature importance distributions.}
As described in \secref{Methods}, our method outputs the parameters of the Dirichlet distribution over the feature importance. 
With the deliberate choice of Dirichlet distribution, we can obtain a closed-form distance metric such as the KL divergence between two BIF's learned Dirichlet distributions. 
We exploit this to study trade-offs between important notions such as explainability, privacy, and fairness. 
In particular, we apply the KL-divergence to describe the level of explainability sacrificed at the cost of increase in privacy of a classifier. We show a similar experiment for the fairness trade-off  in the supplementary materials.

\noindent

%

\begin{figure}[t]
    \centering
    \includegraphics[width=0.8\textwidth]{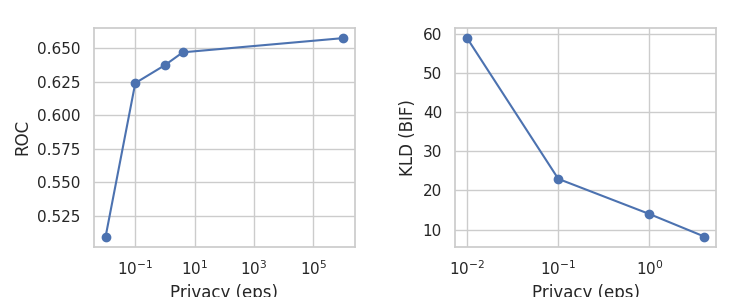}
    \caption{Privacy vs feature importance on Diabetes Readmission data~\citep{rizvi}. The x-axis indicates the privacy level of a classifier (smaller $\epsilon$ means more privacy).
\textbf{Left:} The classification accuracy (ROC) improves as the privacy level decreases.
\textbf{Right:}  KL divergence between the feature importance distribution under the non-private classifier and that under the private classifier at the level that x-axis indicates. 
The feature importance learned by BIF at a stronger privacy (small $\epsilon$) has a larger divergence from the feature importance learned non-privately.
}
    \label{fig:privacy_bar}
    \vspace{-0.3cm}
\end{figure}

We use the Diabetes Readmisison dataset\footnote{We followed the data pre-processing given in https://www.kaggle.com/victoralcimed/diabetes-readmission-through-logistic-regression} \citep{rizvi} (license: CC0 1.0)
to train a private classifier. 
We consider a private classifier using the differentially private stochastic gradient descent (DP-SGD) technique \citep{dp_sgd}, which perturbs the gradients during training to yield a classifier that guarantees a certain level of privacy, that is, it ensures that we cannot recreate the data that the model has been trained on.
\figref{privacy_bar} (Left) shows how the classifier loses the accuracy measured in terms of the area under the curve as we increase the privacy level. Different privacy levels introduce different levels of noise we induced to the gradients during training (the higher $\epsilon$, the smaller the noise level, and $\epsilon=\infty$ corresponds to the non-private classifier). 
%
As the loss of accuracy is a known phenomenon, we aim to show a different effect, namely the impact of noise on the possibility to explain the data in form of the difference between the feature distribution without the noise and that when the noise is present at different levels.

\begin{figure}[t]
    \centering
    \begin{minipage}{0.65\textwidth}
         \includegraphics[trim=60 0 0 0,clip,width=1.0\textwidth]{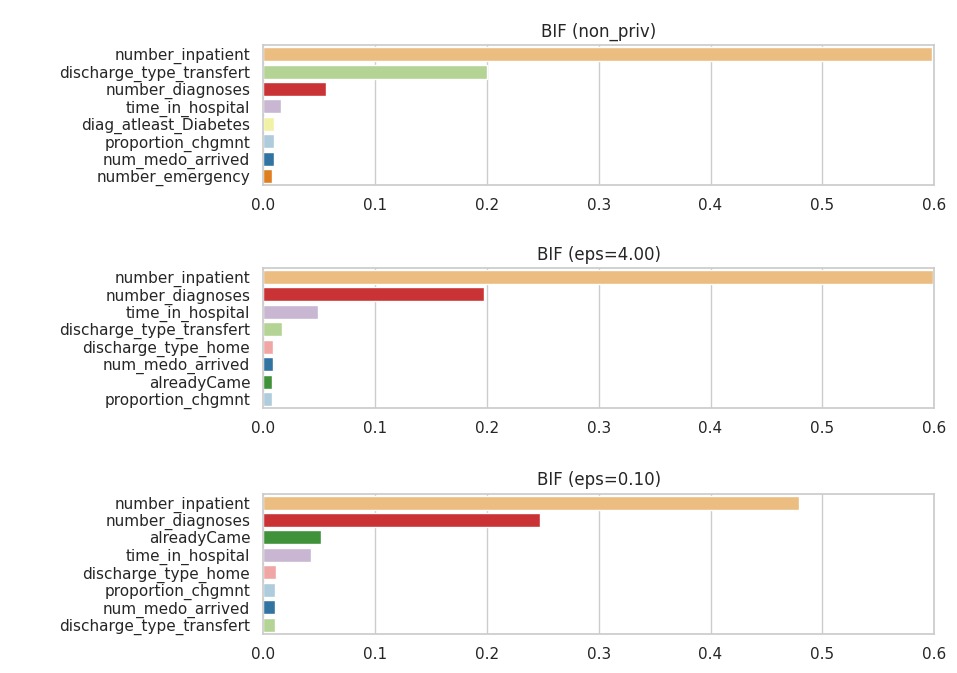}
    \end{minipage}\hfill
    \begin{minipage}{0.05\textwidth}
    \end{minipage}\hfill
     \begin{minipage}{0.3\textwidth}
            \caption{\textbf{
    BIF's learned feature importance}. We present the top $8$ important features to the classifier trained with the Diabetes Readmission data \citep{rizvi} at a different level of privacy. Each feature is color-coded for better visualization.
    Smaller $\epsilon$ indicates higher levels of privacy.
     Relative differences between the features are affected by the increasing privacy (which is reflected in the decrease in KLD in the right \figref{privacy_bar}.).   The most important features in non-private setting remain important even for high levels of privacy, showing a level of robustness for identifying important features.
    }  
     \end{minipage}
    \label{fig:important_features_by_BIF}
    \vspace{-0.5cm}
\end{figure}


The experiment shows two things. Firstly, as \figref{privacy_bar} and the KL divergence chart of \figref{important_features_by_BIF} demonstrate, the relative differences in importance between features decrease as we increase the levels of noise. BIF-tuned probabilities well reflect the intuition that as we increase privacy levels,  the explainability in form of assessing the correct distribution of feature importance decreases. In the Supplementary material, we also include the analysis regarding the INVASE, which shows that this intuition may not be exactly reflected in how features are selected. Secondly as \figref{important_features_by_BIF} shows, even though the relative differences may be obfuscated by the noise, even at high levels of privacy we may distinguish the most relevant features, showing robustness of the feature selection ranking to the noise. 

%% file: table11.tex
\begin{table}[t]
\centering

\begin{tabular}{c|ccc|ccc}
\toprule
& Syn 1 & Syn 2 & Syn 3 & Syn 4 & Syn 5 & Syn 6 \\ \midrule
\rowcolor{Gray}
BIF (global) & $\bm{100}$ & $\bm{100}$ & 85.0 & - & - & - \\
\rowcolor{Gray}
BIF (inst, samp) & $\bm{100}$ & $\bm{100}$ & 93.6 & 81.9 & $\bm{86.0}$ & 85.2 \\
\rowcolor{Gray}
BIF (inst, pe) & $\bm{100}$ & $\bm{100}$ & 82.9 & $\bm{84.0}$ & 79.0 & $\bm{85.6}$ \\
L2X & $\bm{100}$ & $\bm{100}$ & $\bm{95.1}$ & 66.0 & 64.5 & 73.8 \\
INVASE & $\bm{100}$ & $\bm{100}$ & 81.0 & 57.3 & 50.0 & 36.1 \\
SHAP & 98.8 & 98.9 & 93.2 & 59.1 & 59.0 & 49.3 \\
LIME & \textbf{100} & \textbf{100 }& 27.9 & 29.8 & 34.5 & 19.5  \\
\bottomrule
\end{tabular} 

\vspace{0.3cm}
   \captionof{table}{ \textbf{Synthetic datasets} to detect ground truth features. The Syn1-3 datasets consists of a fixed set of  globally invariant important features, while Syn 4-6 consists of varying sets of important features instance-wise. The average over 5 runs is reported. The higher MCC (Matthews correlation coefficient), the better.  In BIF, (\textit{inst, samp}) means instance-wise explanation with sampling, while (\textit{inst, pe}) means that with point estimate. }
    \label{tab:table_synthetic}
    
\end{table}

%% file: table1.tex
\begin{figure*}

\begin{minipage}{.6\textwidth}

\includegraphics[width=0.84\textwidth, trim={0 0 0 1.3cm},clip]{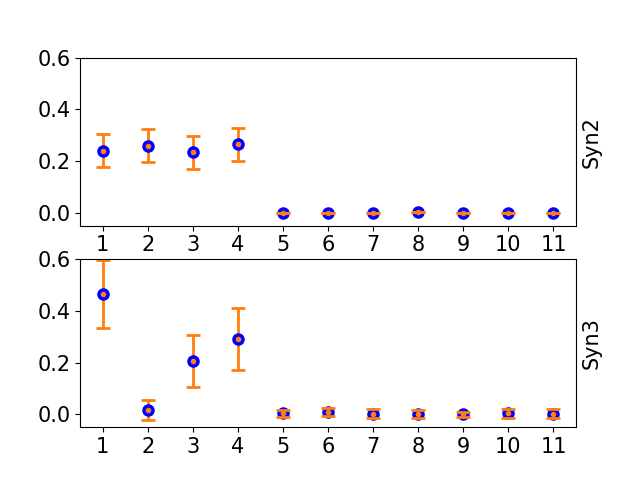}

\end{minipage}
\hspace{0.05cm}
\begin{minipage}{.37\textwidth}
\centering

\caption{The illustration of the importance values and uncertainty learnt by BIF for the two datasets with four important values, Syn2 (Top) and Syn3 (Bottom). The algorithm is appropriately less certain about Syn3 ($\sum \sigma = 0.49$) which consists of features of varied importance (also less accurate) compared to Syn2 ($\sum \sigma = 0.25$).}
\label{fig:means_and_std}
\end{minipage}
\vspace{-0.5cm}
\end{figure*}

%% file: table2.tex
\begin{table*}[t!]
\centering
\begin{tabular}
{p{0.15cm}c | ccc | ccc | ccc }
\toprule
& & & Adult & & & Credit & & & Intrusion & \\
& $k$ & 1 & 3 & 5 & 1 & 3 & 5 & 1 & 3 & 5  \\ \midrule


\rowcolor{Gray}
\cellcolor{white} \parbox[t]{2mm}{\multirow{5}{*}{\rotatebox[origin=c]{90}{Local}}} &

\textbf{BIF} & \textbf{80.0} & \textbf{81.8} & 82,3 &
\textbf{90.5} & \textbf{92.9} & \textbf{94.1} 
 & 81.6 & \textbf{95.8} & 83.6 \\

& L2X
&78.6& 81.7 & \textbf{83.1} 
& 86.5 & 89.1 & 92.8 
 & \textbf{81.9} & 79.0 & 77.4\\
 
& INVASE &
73.5 & 78.6 & 82.1 & 
81.4 & 90.9 & 91.5
 & 70.5 & 76.6 & 45.5 \\

& SHAP 
& 71.8 & 74.8 & 76.9 
& 86.8 & 84.9  & 84.5
& 69.1 & 72.6 & 73.4
\\

& LIME 
& 77.6 & 78.9 & 80.6 
& 85.2 & 87.3 & 94.3 
& 78.0 & 89.5 & \textbf{83.8}
\\

\midrule

\rowcolor{Gray}
\cellcolor{white}
\parbox[t]{2mm}{\multirow{5}{*}{\rotatebox[origin=c]{90}{Global}}} &

\textbf{BIF} & 
\textbf{78.2} & 76.5 & 82.4 & 
\textbf{96.1} & \textbf{94.9} & 94.9 & 
\textbf{82.3} & 82.4 & 82.6  
\\
& L2X & 65.5 & 77.2 & 80 
& 82.6 & 92.2 & 95.2 &
 39.3 & 59.9 & 81.1
\\
& INVASE & 
65.5 & \textbf{82.3 }& 82.4 &
95.5 & 90.7 & 94.6 & 
44.3 & 82.3 & 82.3  
 \\
& SHAP & 76.6 & 79.7 & \textbf{83.1}
& \textbf{96.1} & 94.3 & \textbf{96.4}
& \textbf{82.3} & \textbf{87.1} & \textbf{87.1} 
\\
& LIME & 77.6 & 78.9   & 75.9 
& 92.4 & 88.9  & 92.5
& \textbf{82.3} & 81.3 & 83.4 \\
\bottomrule
\end{tabular}
\vspace{0.3cm}
\caption{\textbf{Tabular datasets}. Classification accuracy as a function of $k$ selected features. \textbf{Up:} For gaining global explainabiilty. Same features are selected for all the datapoints. \textbf{Down}: For gaining local (instance-wise) explainability. A set of $k$ features is selected for each data point separately. }
\label{tab:real}
\vspace{-0.35cm}
\end{table*}